# PREDICTING RESPONSE-FUNCTION RESULTS OF ELECTRICAL/MECHANICAL SYSTEMS THROUGH ARTIFICIAL NEURAL NETWORK

R.C.Gupta*, AnkurAgarwal**, RuchiGupta+, SanjayGupta++

**ABSTRACT:**

In the present paper a newer application of Artificial Neural Network (ANN) has been developed i.e., predicting response-function results of electrical/mechanical system through ANN. This method is specially useful to complex systems for which it is not possible to find the response-function because of complexity of the system. The proposed approach suggests that how 'even without knowing the response-function, the response-function results can be predicted with the use of ANN to the system'.

The steps used are: (i) Depending on the system, the ANN-architecture and the input & output parameters are decided, (ii) Training & test data are generated from simplified circuits and through tactic-superposition of it for complex circuits, (iii) Training the ANN with training data through many cycles and (iv) Test-data are used for predicting the response-function results.

It is found that the proposed novel method for response prediction works satisfactorily. Thus this method could be used specially for complex systems where other methods are unable to tackle it. In this paper the application of ANN is particularly demonstrated to electrical-circuit system but can be applied to other systems too.

\*   Prof.& Head, Mechanical Engg.Dept. Institute of Engg.& Technology, Lucknow, India
\*\* B.Tech. student, Mech.Engg.Dept., Institute of Engg.& Technology, Lucknow, India
\+   Graduate student, Electrical Engg.Dept., Stanford University, California, USA
++ Avesta Computer Services, California, USA

# 1. INTRODUCTION:

Electrical-system ( $L\, d^2q/dt^2 + R\, dq/dt + 1/C\, q = E_0 \sin \omega t$ ) and Mechanical-system ( $m\, d^2x/dt^2 + b\, dx/dt + k\, x = F_0 \sin \omega t$ ) are quite equivalent to each other. So, study of either system is enough as its results can be applied to other system as well. In the present paper the concentration has been for study of the response in electrical-system, to be more specific to electrical circuits. Electrical-circuit input elements are resistances(R), inductances(L), capacitances(C) and voltages(V or E). The output response parameters are current(I or i) and phase($\varphi$) in an element.

The electrical circuit may contain in addition to voltages, several but only-resistances (R) or may also-contain inductances (L) and capacitances (C) . Analysis for the 'only-resistances' circuits for output response (current i in an element) prediction/calculation could be made with the help of methods such as Kirchoff's law, Thevevnin's theorem, Norton's theorem etc.[1,2], but as the number of loops & its elements increases complexity of the analysis also increases. For L,C, R circuits; even the simple LCR circuit containing a few loops are difficult to analyze for output response (I , $\varphi$) prediction in an element, but as the number of loops increases complexity mounts to such an extent that it is almost impossible to find response in complex LCR circuits. Although for large scale circuit design & performance, computer-aided circuit-simulator may be used but has limitations from technological & economic point of view.

Here comes Artificial Neural Network (ANN) to rescue to tackle this difficult/impossible situation i.e., for large number of large-scale complex circuits. The approach presented in the paper introduces a novel application of ANN to electrical-circuit; which explains that 'even without knowing the response-function, it is possible to predict the desired response-function results'.

## 2. ARTIFICIAL NEURAL NETWORK FOR ELECTRICAL CIRCUIT :

Artificial Neural Network (ANN) modeled on human brain finds output from inputs {3,4}. The feed-forward back-propagation supervised learning ANN model is post popular and is commonly used. In fact ANN is the modeled mathematical-brain to predict output from given inputs, if trained earlier with human data. ANN is more suitable specially if the relationship between input & output is not-known or not-clear or apparently very-complex and un-understandable/un-deducible. The complex electrical circuit with many loops specially containing several L,C,R are such situation; and warrants use of ANN to predict output (current I or i , phase $\varphi$ in an element) for given inputs (voltages V or E, resistance-elements L,C,R).

**2.1 Electrical-Circuit and the ANN-Architecture for it**

For better understanding and gradual up-gradation of complexity, the electrical circuits are classified in two categories with three types of loops in each category, as follows:

    Category:    1. Circuits with resistances-only
                     2. Circuits with L,C,R

    Loop:        a. One loop circuit
                     b. Two loop circuit
                     c. Four loop circuit

Thus a total of six types of circuits are considered. The simplest(1.a) among that could mean Ohm's law (i = V/R) type circuit (Fig1); while the four loop circuit with several L,C,R (2.c) would be too complex / difficult to analyze (Fig.2) to the level of indeterminacy. It is shown in the paper that ANN can be used to predict output from inputs in all situations ranging from simple to difficult circuits.

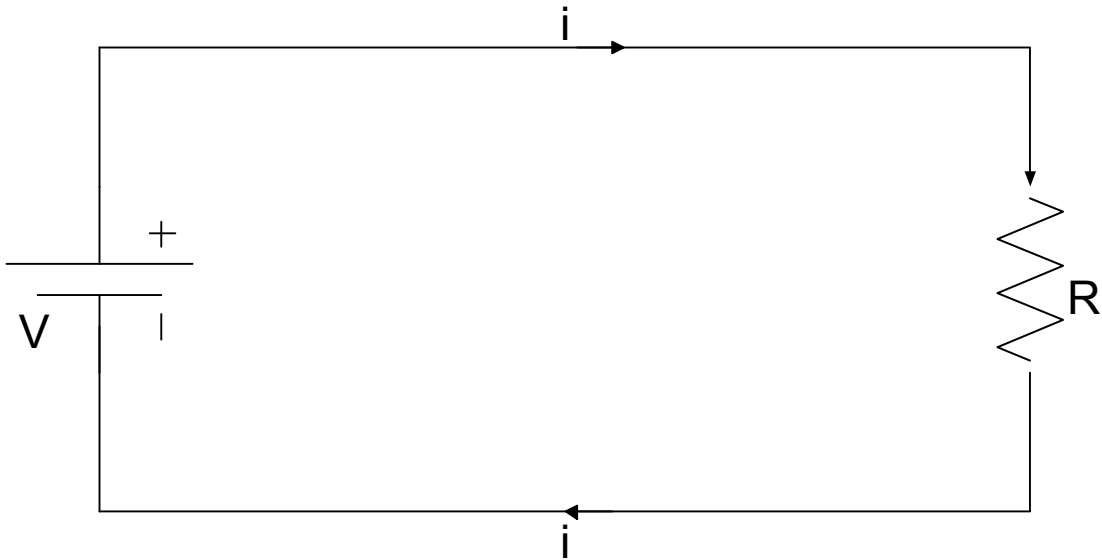

**Figure 1. Simple Ohm's law type circuit (Category-Loop : 1-a )**

Artificial Neural Network (ANN) feed-forward with back-propagation supervised-learning is to be used. The ANN architecture strategy is decided to have usually four layers (one input, one output, two hidden layer). The input parameters are impedance elements ($R, X_l, X_c$) & voltages (V or E) and output being current (i & φ). For clarity of details the listing of parameters are given in tabular form.

| Category & loop type | Input parameters Impedance elements | Voltages | Hidden layers & neurons in each | | output parameters current | phase |
|---|---|---|---|---|---|---|
| 1.a | R | V | 1 | 3 | i | - |
| 1.b | $R_1, R_2 \ldots R_7$ | $E_1, E_2 \ldots E_7$ | 2 | 8 | $i_6$ | - |
| 1.c | $R_1, R_2 \ldots R_{12}$ | $E_1, E_2 \ldots E_{12}$ | 2 | 16 | $i_6$ | - |
| 2.a | $X_l, X_c, R$ | E | 2 | 3 | i | $\varphi$ |
| 2.b | $R_1, X_{c1}, X_{l1}, \ldots R_7, X_{c7}, X_{l7}$ | E | 2 | 15 | $i_6$ | $\varphi$ |
| 2.c | $R_1, X_{c1}, X_{l1} \ldots R_{12}, X_{c12}, X_{l12}$ | $E_1, E_2 \ldots E_{12}$ | 2 | 25 | $i_6$ | $\varphi$ |

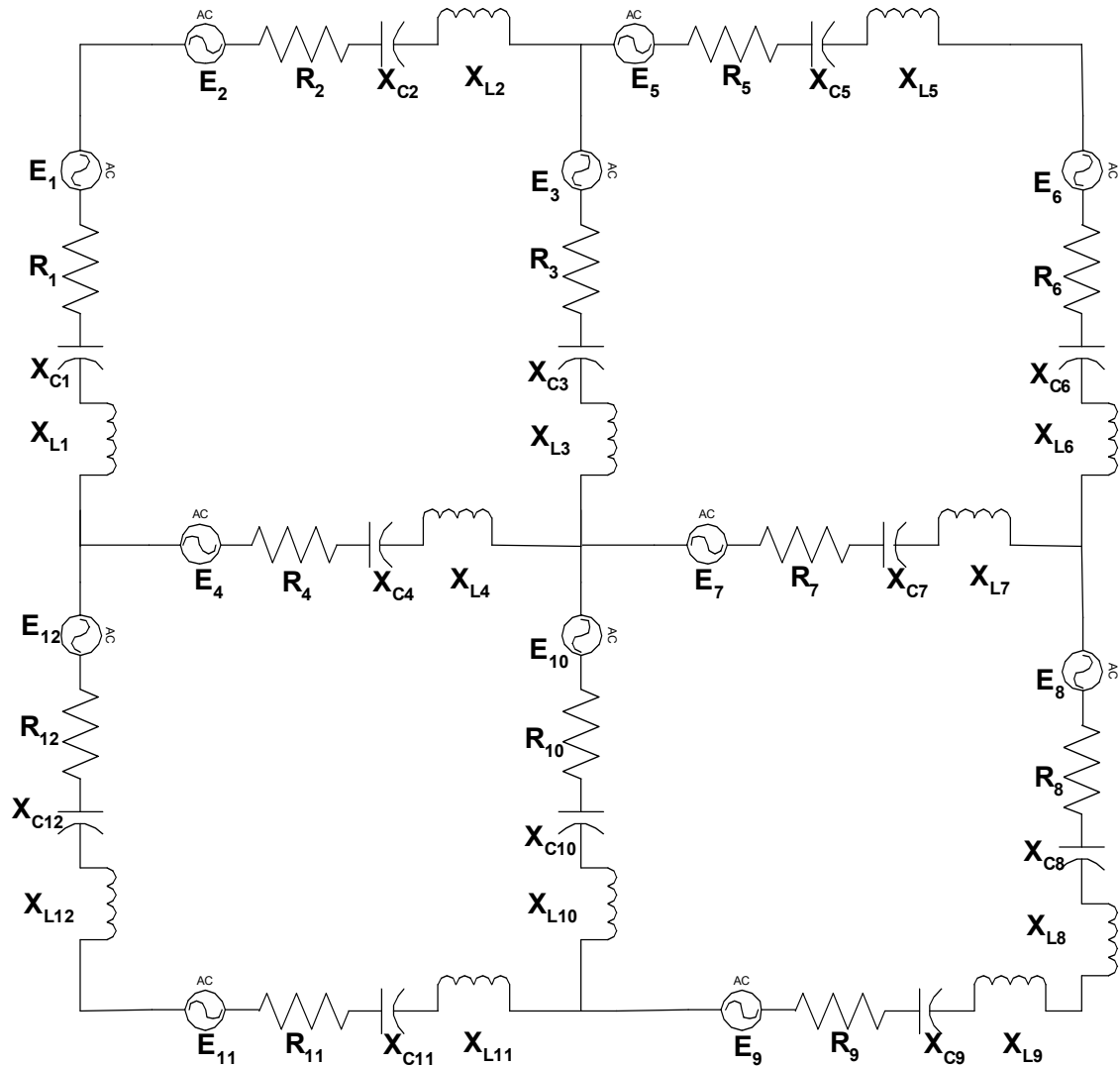

**Figure 2. Complex LCR Circuit (Category-Loop 2.c)**

**2.2 Training & Test Data**

There is no problem for generating training & test data for simple circuits such as that of 1.a,1.b and 2.a. The training for more complex circuits as 1.c and 2.b &2.c (since these are difficult to analyze) are generated [5] typically by symmetry, superposition and other tactic considerations. A few test data are prepared out of a few training data in which the known-output is suppressed and it is to be predicted through the trained-ANN for comparison.

Modification of data is to be made to avoid zeros & infinities for better functioning of ANN. The open-circuit-path with infinite resistance is taken as path with very high resistance say 100 ohm and that short-circuit –path with zero resistance is taken as path with very low resistance say 0.1 ohm. Similarly, no or zero voltage is replaced with 0.1 volt.

Normalization of training & test data is done to bring the data preferably in 0 to 1 range. This is usual practice followed for better performance & results. Normalization is done in a way to give almost equal importance to all set of data. The normalization convention for each type of circuit is done such that for impedance elements $R_{norm}=R/R_{max}$, $X_{l norm}=X_l/X_{l max}$, $X_{c norm}=X_c/X_{c max}$, for voltage V or $E_{norm}=V$ or $E/ V$ or $E_{max}$, for current i or $I_{norm}=i$ or $I /(0.1 E_{max})$, for phase $\varphi_{norm}=(\varphi+180)/360$. Sometimes even after normalization 0.01 is added to all data to get rid of zero if any and to deliberatory introduce some noise into the data for better learning on tough road.

**3. ARTIFICIAL NEURAL NETWORK FOR OHMS LAW**

The aim of this paper is the use of Artificial Neural Network (ANN) to predict the response specially for complex electrical circuits. But anything that glitters is not gold ! The use of ANN even in most simple (Fig.1) one loop resistance circuit (for Ohm's law $I = V/R$) faces problem in the beginning. In ANN the output i is weighted sum of inputs V & R but with sigmoidal modification function to act upon it. It is intriguing 'how such a 'division' function (i =V/R) behavior could be predicted as 'weighted sum' of V & R. But it is found [5] that the 'properly trained' ANN works for Ohm's law and it works well ! It is possibly due to the sigmoidal function which acts upon the weighted sum and possibly due to inter-action between the neurons in the hidden layer. This leads to a confidence that ANN can work for response prediction in complex circuits satisfactorily.

**4. PREDICTING RESPONSE-FUNCTION RESULTS OF ELECTRICAL SYSTEM THROUGH ANN :**

Each of the six circuits 1.a,b,c and 2.a,b.c are dealt with separately one by one. The ANN for the particular type of circuit is trained with previously generated training data. It may be noted that since the data is limited in number, the ANN must be trained for more number of cycles; analogically, as if a doctor(ANN) being trained through evaluating a limited number of

patients(data) more often(cycles). For example; for 50 sets of training data, at least 500 cycles training works satisfactorily, that means the doctor examining each patient at least 10 times.

The ANN is trained with training data using feed-forward back-propagation ANN-software (software usually easily available with books, e.g., C++ Neural Network & Fuzzy Logic by Rao & Rao). Initially the program assigns connecting weights randomly which are adjusted/modified with the back-propagation algorithm as the training proceeds. After initial training, there is provision to remember (assign) the previous-adjusted-weights for re-adjustment in subsequent training; and this is better as it retains the previous learning for further learning.

After 'proper training', it is possible to predict the output response from input data using the test data. There is in fact no need to know the exact output response function (output as a function of inputs). The trained-ANN suffice the purpose as it is able to predict the output response function results, even without knowing the response-function itself. The evaluation with the test data shows [5] satisfactory predictions in all circuit types & loops.

## 5. ARTIFICIAL NEURAL NETWORK USE EXTENDED TO SIMPLE ELECTRONIC -CIRCUIT:

Encouraged with the success of use of ANN to Electrical-circuit it is tempting to see if ANN could be applied to Electronic-circuits too. Electrical-circuits contain elements such as resistors, inductors & capacitors whereas usually electronic circuit contain elements such as resistors, diodes & transistors. Mixed complex circuits may include all these elements in modern devices including IC and LSI.

It was, however, decided to check for simple amplifier electronic-circuit which is similar to two-loop electrical-circuit. Though similar in some sense, but the electrical-circuit & electrical circuit (Fig.3) are quite different. In electrical-circuit Kirchoff's law is used in each loop to find output current ($i_6$) whereas for electronic-circuit only ohm's is used in each loop (assuming negligibly-small base-current). The training and test data for both the circuits (electrical & electronic) are generated, the method and thus the values for both type of circuits are quite different. ANN architecture for both type of circuits are same, fixing the middle resistance $R_3=1$ the inputs being $R_1$, $R_6$ & V and output as $i_6$ and with two hidden layers with three neurons in each.

After training separate ANNs one for the electrical-circuit and the other for electronic-circuit, it is found [5] that both the ANNs work satisfactorily well. Thus it is concluded that ANN not only works well for simple to complex electrical-circuits but it also works well at least for simple electronic-circuit.

With these successes it is anticipated that ANN would also work for complex electronic-circuits and also for mixed circuits containing both electrical & electronic elements . The authors expect that this approach through ANN for predicting response-function results (even without knowing the response-function itself) for electrical & electronic systems will grow further and find wider application for computer-aided circuit design.

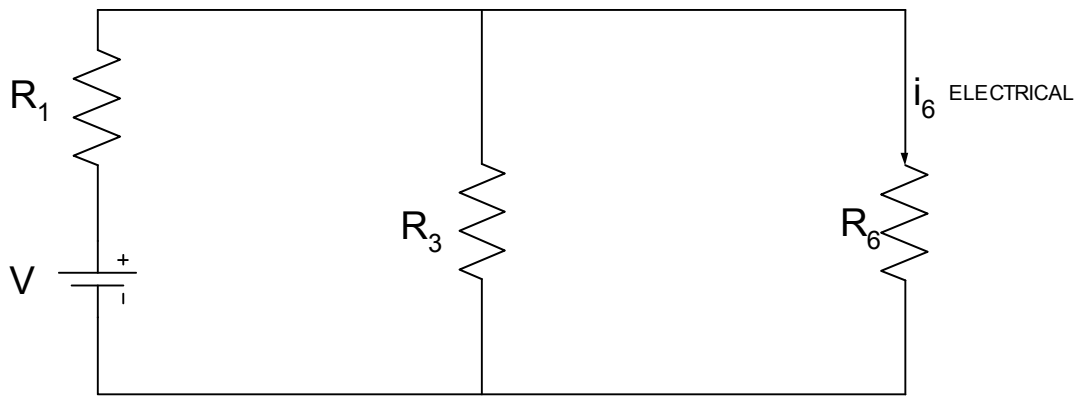

(a) Simple Electrical Circuit

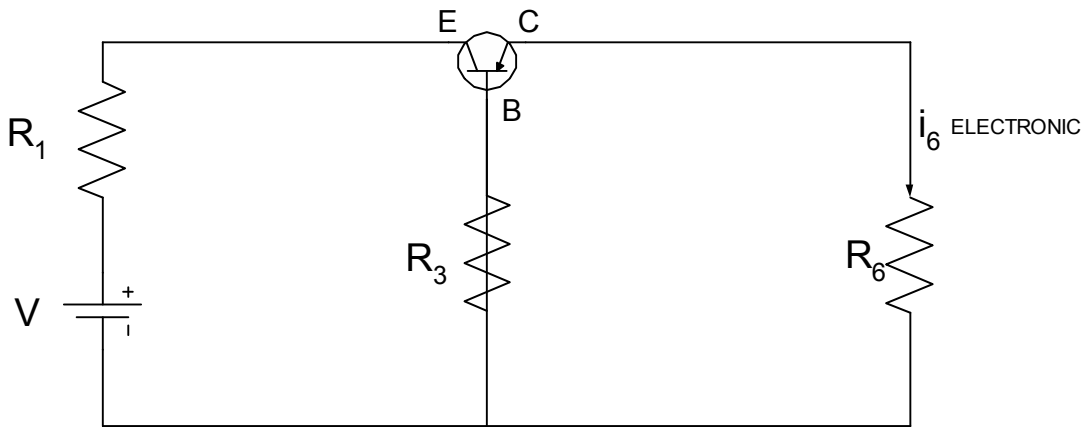

(b) Simple Electronic Circuit

**Figure 3. Electrical and Electronic Circuits**

## 6. POSSIBLE FUTURE APPLICATIONS OF THE USE OF ANN TO ELECTRICAL & ELECTRONIC SYSTEMS

The present paper introduces the novel technique of use of ANN for predicting the response in electrical & electronic circuits. The future applications of it could be its use for more complex electrical & electronic circuits and for mixed circuits containing both electrical & electronic elements & devices. The future application of it would be for computer-aided-design of circuits. In future the method presented in the paper could be extended to predict response in integrated circuits (ICs) and possibly to LSI and VLSI.

The authors suggest addition of a pre-processor and a post-processor to the ANN-software. The pre-processor would be small computer-program to generate large number of various possible circuit-types with varying values of circuit-elements at quantum intervals; the pre-processor also set to generate training & test data and to modify & normalize it accordingly for a few circuits only. So, one can have large number of possible circuits along with the training & test data for few circuits to train the ANN. Now, this trained-ANN could be used to predict the output response for large number of various possible circuit types; finding output response for 'large number' of various possible circuits is practically not possible otherwise either experimentally or theoretically. The post-processor is a computer program to select or pick the good or better ones; although the trained-ANN will find response for all of the large number of circuits but post-processor is to filter/suppress the results of poor circuits and only the results of good or better ones will be reported to the user. In a way, the authors suggest a novel 'virtual instrumentation with ANN' for computer-aided circuit design.

## 7. CONCLUSIONS:

Electrical & Mechanical systems are equivalent, and study of one system is enough to understand the other. It is shown in the paper that complex electrical system (which are otherwise almost impossible to analyze) could be studied for response with Artificial Neural Network (ANN). It is shown that how even without knowing the response-function, the response-function results in electrical-circuits can be predicted through ANN. Predictions are satisfactory and encouraging. An extension of this approach to electronic-circuit also works.

The novel method to find response in electrical/electronic/mechanical through ANN suggested in this paper is not the final word. In fact it is a beginning of a new technique/approach to tackle complex problems and for computer-aided circuit design.

## ACKNOWLEDGEMENT:

Thanks are due to Lalit Kumar and Sunil Kumar Prabhakar for being with Ankur Agrawal for assistance.

## REFERENCES:

1. A.E.Fitzgerald, David E Higginbotham and Arvin Grabel, 'Basic Electrical Engineering', McGraw Hill Publications, 2000.
2. B.L.Theraja and A.K.Theraja, 'A Textbook of Electrical Technology'' S.Chand Publications., 1997.
3. Bart Kasko, 'Neural Network and Fuzzy System', McGraw Hill, 1981.
4. V.B.Rao and H.V.Rao,' C++ Neural Network and Fuzzy Logic', BPB Publications, 2000.
5. 'Predicting Response-Function Results for Electrical Mechanical System through Artificial Neural Network', Graduate Project lead by Ankur Agrawal under inspiration & supervision of Dr.R.C.Gupta.